# Feedback Regulated Opto-Mechanical Soft Robotic Actuators

**Authors**

Jianfeng Yang,[1] Haotian Pi,[2] Zixuan Deng,[1] Hongshuang Guo,[1] Wan Shou,[3] Hang Zhang,[2] Hao Zeng[1*]

**Affiliations**

[1] Light Robots, Faculty of Engineering and Natural Sciences, Tampere University, P.O. Box 541, FI-33101 Tampere, Finland

[2] Life-inspired Soft Matter, Department of Applied Physics, Aalto University, P.O. Box 15100, FI 02150 Espoo, Finland

[3] Department of Mechanical Engineering, University of Arkansas, Fayetteville, AR 72701, USA

*Correspondence to hao.zeng@tuni.fi

**Abstract**

Natural organisms can convert environmental stimuli into sensory feedback to regulate their body and realize active adaptivity. However, realizing such a feedback-regulation mechanism in synthetic material systems remains a grand challenge. It is believed that achieving complex feedback mechanisms in responsive materials will pave the way toward autonomous, intelligent structure and actuation without complex electronics. Inspired by living systems, we report a general principle to design and construct such feedback loops in light-responsive materials. Specifically, we design a baffle-actuator mechanism to incorporate programmed feedback into the opto-mechanical responsiveness. By simply addressing the baffle position with respect to the incident light beam, positive and negative feedback are programmed. We demonstrate the transformation of a light-bending strip into a switcher, where the intensity of light determines the energy barrier under positive feedback, realizing multi-stable shape-morphing. By leveraging the negative feedback and associated homeostasis, we demonstrate two soft robots, i.e., a locomotor and a swimmer. Furthermore, we unveil the ubiquity of feedback in light-responsive materials, which provides new insight into self-regulated robotic matters.

**Teaser**

Positive and negative photomechanical feedback is readily programmed in a soft actuator.

# MAIN TEXT

## Introduction

Responsive materials can sense, respond to, and interact with their external environment, unlike traditional static material, which faces difficulties in altering their inherent properties. Examples include shape memory polymers (*1, 2*), hydrogels (*3, 4*) and liquid crystal elastomers (LCE) (*5-7*), which can dynamically change their shapes upon external stimuli (light, heat, moisture, *etc.*). Currently, the paradigm is shifting towards developing intelligent materials systems through a highly architected combination of responsive materials. In such a system, a series of autonomous dynamic behaviors reveals the unseen and inspiring potential that differs from that of single responsive materials (*8, 9*). This behavior cannot be achieved merely by stacking responsive materials together; it requires more precise identification of key points that are taken from a bioinspiration perspective. Any biological system is a typical and complex autonomous system that steadily absorbs energy or supplies from the external environment and dissipates a portion to its surroundings to maintain fundamental physiological processes and confront various stochastic events (*10, 11*). Signal transduction occurs at both intracellular and intercellular levels, coordinating stimulus-response processes across different temporal and spatial scales, helping individual cells sense and respond to their environment (*12, 13*). Feedback regulation is a crucial mechanism ensuring effective signal transduction and adaptive responsiveness (*14, 15*). Therefore, constructing feedback loops within responsive materials could be an effective way to achieve autonomy and dynamic behaviors similar to those of living organisms.

In biological systems, both positive and negative feedback serve as cornerstones for self-sustaining living functions, *e.g.,* the cell cycle (*16*), signal cascades (*14*), enzyme reactions (*17*), and cell differentiation (*18*). Typically, positive feedback is one of the mechanisms of epigenetic inheritance, with the perpetuation accompanying cell differentiation being considered an irreversible bistable phenomenon resulting from strong positive feedback (*14*). Negative feedback, for example, controls the reaction rate of substrates in various enzymatic reactions to increase the stability of a multi-step reaction, which often manifests as various rhythmic oscillations in biological systems (*19*). Signal transduction generally involves multiple steps, with feedback occurring when a downstream step affects an earlier one (*20*). Positive feedback amplifies actions, while negative feedback inhibits upstream signals. In synthetic systems, the interaction between the external environment and responsive materials is typically unidirectional. This means that after initial transience, the system's response corresponds to the steady state of the external input stimulus, without further interaction with the input. To enable non-unidirectional interactions with the environment, it is necessary to build feedback loops through rational structural design based on responsive materials. Recently, there have been examples of integrating feedback loops in responsive matters that showcase advanced stimulus-response behaviors similar to biological behaviors, such as phototropism (*21*), homeostasis (*22*), and sensation (*10*). However, there is no general approach available for the rational construction of positive and negative feedback loops through structural design in light-responsive materials.

Here, we present the design of an opto-mechanical system by using a soft actuator, in which both positive and negative feedback can be obtained. Constructing such a feedback mechanism enables active interactions of a soft robotic structure with external illumination. We present dynamic multi-stable states separated by energy barriers via positive feedback, where the memory is retained even after the light is withdrawn. The system constructed with negative feedback on one hand utilizes part of the input energy to maintain its homeostasis and on the other hand dissipates energy to perform work in the external environment, enabling locomotion functions like walking and swimming. Furthermore, we elucidate the prevalence of feedback mechanisms in light-responsive materials with different deformation modes.

## Results

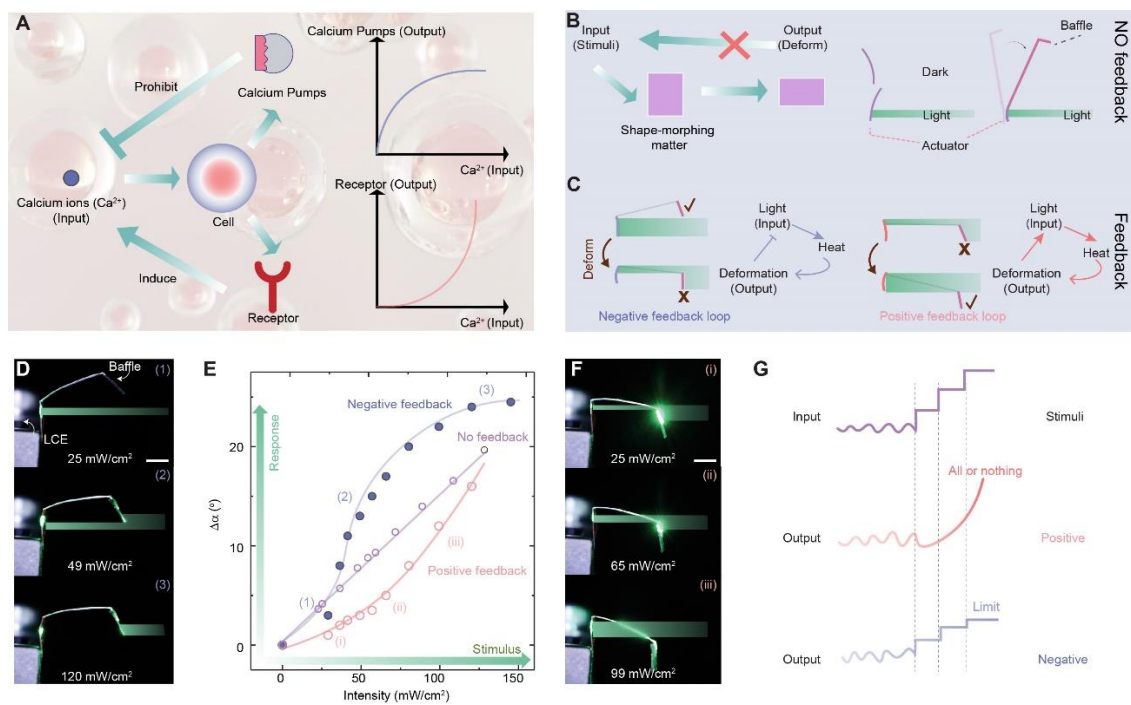

**Figure 1. Feedback actuator systems.**
(A) Positive feedback and negative feedback mechanisms involved in cell system. (B) Left, schematics of shape-morphing soft matters deforming (output) upon external stimuli (input); Right, schematics of a photothermal strip that reversibly deforms upon light irradiation. (C) Schematics of negative feedback (left) and positive feedback (right) designs of the baffle-actuator systems. (D) Photographs of negative feedback baffle-actuator systems under varying light intensity irradiation. (E) Variation of Δα in the photothermal strip with increasing light intensity under no feedback (purple), negative feedback (blue), and positive feedback (red). (F) Photographs of positive feedback baffle-actuator systems under varying light intensity irradiation. (G) The relationship between input and output of positive feedback and negative feedback mechanisms. Light: 532 nm. All scale bars are 5 mm.

The distinguishing features of life predominantly hinge on dynamic non-equilibrium states, characterized by continual energy dissipation and orchestrated through feedback loops. For example, calcium serves as a crucial secondary messenger in cells, meaning that fluctuations in its concentration can rapidly activate or deactivate various cellular functions. The regulation of calcium levels is governed by complex feedback systems that ensure proper cellular function. As shown in Fig. 1A, when intracellular calcium levels increase due to signaling events such as neurotransmitter release or muscle contraction, calcium pumps are activated to expel excess calcium from the cytoplasm (negative feedback) (*23*). As calcium levels rise, these pumps work more intensively to restore balance. Additionally, even small amounts of calcium entering the cytoplasm can bind to receptors on the sarcoplasmic reticulum, prompting the release of stored $Ca^{2+}$ (positive feedback) (*24*). This amplifies the calcium signal, enabling neurons to respond effectively to stimuli, such as initiating neurotransmitter release or activating downstream processes like gene expression. In general, the feedback mechanism in biology is a regulatory process in which the output of the pathway will positively or negatively affect the early process of the pathway, causing subsequent adjustments in the system's behavior. This concept inspired us to investigate the counterpart in shape-morphing soft matters, as illustrated in Fig. 1B (left), where the external stimulus acts as inputs to the system, causing corresponding deformations in the material as outputs. Unfortunately, current stimuli-responsive material research lacks the consideration of feedback mechanisms that associate input and output signals.

Here, we take the most general example to illustrate the feedback concept with photothermally deformable polymer (*25, 26*). A strip of sample is characterized by the reversible shape-change (bending) upon an optical field stimulus, as schematically shown in Fig. 1B (right). The degree of deformation at the steady state – when light-induced heat is fully dissipated by the heat loss to the air and the motion ceases – is expected to depend on the incident light intensity. In Fig. 1B right, we showcase a light-bending strip actuator made of liquid crystalline elastomer (LCE) that undergoes deformation with the increase of light intensity. The LCE strips incorporated with dye (*27*) present that the deformation is driven by photothermal absorption (*28, 29*) (fig. S1). Further details regarding the fabrication and mechanical properties of LCE strips can be found in Materials and Methods and figs. S2 and S3. The change in bending angle $\Delta\alpha$ increases linearly along the input light intensity within a certain intensity range (Fig. 1E, purple circle, and fig. S4). This range refers to the intensity below the threshold where the light induces the self-shadowing effect (*30*) (where the material starts blocking the incident light). The linear relationship explicated in Fig. 1E indicates the absence of a feedback mechanism. This stimulus-response relationship can apply to actuation systems with attachments, in which the deformation (output) does not influence the light beam (input). It can also apply to other actuator systems, independent of their chemical nature. Therefore, the design here discussed only deals with opto-mechanics based on physical principles, without any specific chemistry requirement for materials.

Fig. 1C explicates two system designs embedded with nonlinear feedback behavior, wherein the position of the attachments, *i.e.,* the baffles, has different impacts on the material's absorption of the incoming light field, indicating the presence of a feedback mechanism. The baffle, made of aluminum foil, is capable of completely blocking the incident light. Details of the sample preparation can be found in Materials and Methods and fig. S5. In the first case, the baffle is placed on the top of the light beam, initially allowing complete incidence of the excitation beam. As the actuator absorbs the light and deforms, the baffle deflects downward, beginning to block the light field (schematic drawing on the left of Fig. 1C, optical photos in Fig. 1D). The consequent reduction in light delivery serves as the negative feedback to the light absorption, which in turn reduces the photo-mechanical deformation. In other words, the system requires progressively higher light intensity input to achieve a larger bending angle. Consequently, the bending angle saturates with an increase in the light intensity and reaches a plateau (blue dots in Fig. 1E). The second case, schematically shown on the right of Fig. 1C, involves setting the baffle at an initial position that can block most of the light beam, leaving only a small portion of the light to illuminate the actuator. Once the actuator bends, the baffle opens more space for light to propagate to the LCE material, resulting in higher energy delivery to the deformable materials (see optical photos in Fig. 1F). This positive feedback results in an enhanced deformation along increase in incident light intensity, yielding an angle-intensity curve with exponential growth (Fig. 1E, red circle). These two types of light-responsive curves are recorded in samples with different dimensions (fig. S6). The pronounced nonlinearity in the photo-mechanics of such baffle-actuator systems is induced by the feedback regulation, as it is distinct from the linear curve obtained from the construct with an identical LCE strip but without optical feedback (Fig. 1E, purple circle). Usually, nonlinear behavior is exhibited only when the LCE strip without optical feedback begins to soften as it approaches the phase transition temperature under intense light. In contrast, dynamically feedback-induced nonlinear behavior does not require alteration of the material's inherent properties.

The feedback mechanism has inherent similarity with various biological processes and, therefore can also be described by using the motifs that are inspired by biologists, as shown in Fig. 1A. Positive feedback is typically associated with reversible or irreversible bistable state transition, see schematic drawing in Fig. 1G. By reinforcing deviation, positive feedback helps the system transition from an unstable nearby state to another stable state, as will be specifically discussed in Figure 2. Here, light → heat → deformation → light (absorption), where "→" stands for "induce". Negative feedback is nearly omnipresent in known signaling pathways, characterized by sequential regulatory steps that inversely feed back the output signal to the input. These mechanisms are designed to counteract deviations from the set point or desired state, thereby curtailing excessive deviations and contributing to the maintenance of stability and homeostasis (Figure 3). Here, light → heat → deformation —| light (absorption), where "—|" stands for "prohibit". In the following, we will illustrate that specific arrangements between baffle and actuator positions can induce different functions by positive and negative feedback, respectively.

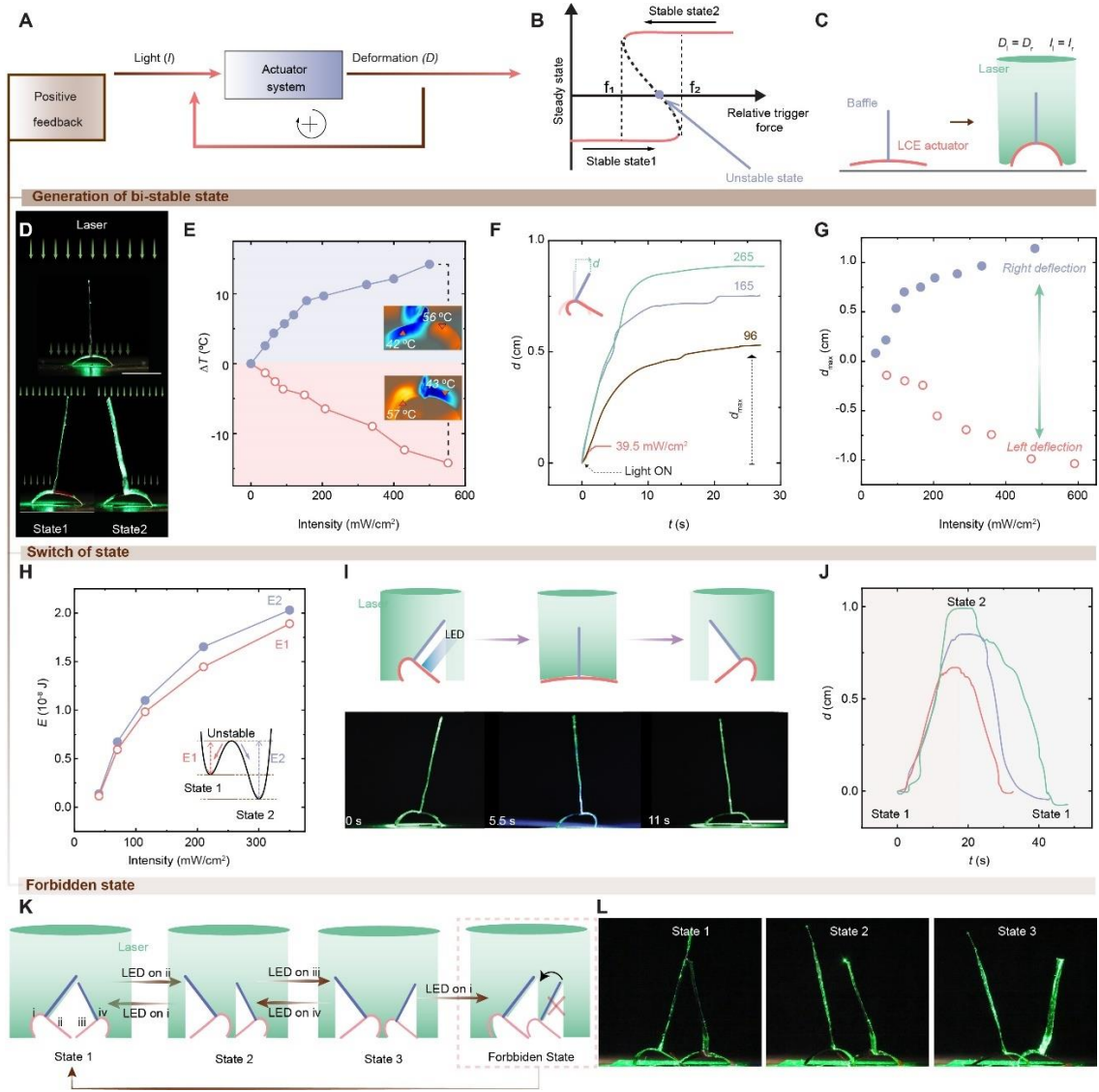

**Figure 2. Positive feedback in the light-response actuator.**
(A) The schematics of the positive feedback mechanism in the light-response actuator. (B) The schematics of bistable state induced by positive feedback. (C) The schematics of a baffle-actuator system with positive feedback. Here, $D_l$ and $D_r$ refer to the deformation of the left and right sides of the actuator, while $I_l$ and $I_r$ denote the incident light energy absorbed by the left and right sides of the actuator, respectively. (D) Photographs of a positive feedback actuator evolving from an unstable to a stable state. Light fuel intensity from the top: 245 mW/cm$^2$. (E) The change in the temperature difference $\Delta T$ between the left and right sides of the light-response actuator under varying light intensity irradiation. Inset: the corresponding infrared images of the light-response actuator. (F) The kinetics of the horizontal displacement $d$ of the baffle tip upon different light fuel intensities. (G) The maximum displacement $d_{max}$ upon different light fuel intensities. (H) The change in energy barrier between two stable states as a function of laser power. Inset: the schematic of the transition of the bistable actuator from state 1 to state 2 and vice versa with respect to free energy. (I) The schematics (top) and photos (bottom) of the switching of state triggered by external light illumination from an LED source. Light fuel intensity: 170 mW/cm$^2$. LED trigger intensity: 270 mW/cm$^2$. LED trigger irradiation direction, from right to left. (J) The kinetics of reversible switching using the identical LED trigger (230 mW/cm$^2$), upon different light fuel intensities. Light fuel excitation: 96 mW/cm$^2$ (red), 120 mW/cm$^2$ (blue) and 150 mW/cm$^2$ (green). (K) The schematics of the switching between three stable states. (L) The switching of three states triggered by external light illumination (an LED source). Light fuel intensity: 160 mW/cm$^2$. LED trigger intensity: 270

mW/cm$^2$. In all cases, the wavelength of light fuel: 532 nm; the wavelength of LED trigger: 460 nm. Scale bars: 1 cm.

Fig. 2A illustrates the positive feedback mechanism of the light-responsive actuator. Light induces deformation of the actuator, which, in turn, amplifies the received light illumination. Consequently, the system moves continuously in a certain direction. This mechanism forms the basis for bistability, as conceptually shown in Fig. 2B. Initially, the system is observed in a stabilized state, then an external suprathreshold disturbance can assist the system in transitioning from one state to another, driven by positive feedback. In our study, a positive feedback-controlled actuator can be designed by fixing a baffle at the center of an LCE strip capable of bending upon light excitation, as schematically shown in Fig. 2C and fig. S7. Ideally, the vertically positioned baffle separates the actuator into two halves, dividing the incident light energy equally onto two sides of the material. As a result, the actuator bends downward, raising the center of mass (Fig. 2C, right), thereby producing two equally deformed structures ($D_l = D_r$, due to $I_l = I_r$). It is worth noting that such a configuration is unstable. Assuming a small fluctuation from the environment causes a slight displacement on the left side, it results in a more pronounced deflection of the baffle to the right, resulting in an increase in temperature on the left side and a decrease on the right. This causes greater deformation on the left, blocking more light from reaching the right side ($I_l > I_r$), and redirecting more light toward the left (state 1 in Fig. 2D, and movie S1). This positive feedback loop ($I_l \rightarrow D_l \rightarrow I_l$) drives the structure to deform continuously until the baffle reaches the right-most position where the left part of the actuator arrives at its maximum deformation. At this point, the temperature difference $\Delta T$ between the left and right sides reaches its maximum, of which value increases with the rise in incident light intensity (Fig. 2E).

We quantify the system's shape-morphing by measuring the horizontal displacement distance $d$ of the baffle tip (Fig. 2F) and record its kinetics by switching on the light fuel at various intensities. Fig. 2F shows that the baffle distance increases logarithmically along with excitation time with a similar decay time of about 10 s. The maximum distance, $d_{max}$, which indicates the location at its mechanically stable state, also increases with the input intensity (Fig. 2G). The symmetry of the structure gives rise to two stable configurations for the same system – the baffle can either deflect to the left or right. More kinetic details about the bi-directional morphing can be found in fig. S7.

Upon external triggering that overcomes the energy barrier, the system can transition from one stable configuration to another. The energy barriers (from state 1 to state 2 and and vice versa) are determined by the light intensity, as shown in Fig. 2H, and increase with the light intensity (detail about the energy barriers estimation, see fig. S8 and Materials and Methods). As illustrated by Fig. 2I, fig. S9 and movie S2, both mechanical and optical triggers can induce state switching – the tip deflection changes from the right side (state 1) to the left (state 2). The corresponding switching under optical triggers in the opposite direction is shown in fig. S10. When a mechanical trigger is applied with insufficient intensity, regardless of duration or direction, the system will revert to its original steady state once the trigger is released. However, a state transition will occur when the trigger's strength exceeds the energy barrier. To quantify the sensation of the system to the external stimulus, we employ LED illumination as a contactless optical trigger to test the switching property. When a triggering light with sufficient long duration and high intensity is applied to the less deformed actuator segment, the structure undergoes mechanical switching. The kinetics of switching exhibit light intensity dependence, as shown in Fig. 2J. This is ascribed to the deterministic role of the light fuel intensity in the energy barrier between two mechanically stable states, driven by the positive feedback regulation. In other words, a higher light fuel intensity dictates a greater energy barrier (Fig. 2H), requiring a higher light trigger intensity to switch the states. A longer triggering time under the same light trigger strength is required when a higher intensity of the light fuel is applied. The intensity thresholds of LED light triggers for switching states under various light fuel intensities and trigger durations are summarized in fig. S11.

In general, a multi-stable actuator can be created by connecting two or more bistable units in series, which can yield $2^N$ possible stable configurations, where N represents the number of bistable units (*31*) (fig. S12). However, in our case, unlike N-discrete individuals without interconnections, the multi-stable model is constructed by establishing dynamic interconnections between units that involve a feedback mechanism. As a result of the varying illumination on each unit, adjacent unit(s) exert mutual influence, leading to an overlap of states and the formation of forbidden states. This reduces the total number of possible stable configurations to fewer than $2^N$. Fig. 2k illustrates the scenario when two bistable units are linked by a light field from the top, and the existence of only three stable configurations. Specifically, as shown in Fig. 2L, when the baffle of the left unit is positioned to the left, the right unit's baffle can be optically switched between right and left directions, resulting in two stable configurations (States 2 and 3). Conversely, if the left unit's baffle is switched to the right (shining the LED on the i part), it blocks light illumination from reaching the left side of the right unit, causing the right-side baffle to stabilize its position only on the left side (State 1). The shadowing between two units prevents one typical configuration of the system, called the forbidden state, as shown in Fig. 2K. The transition kinetics are depicted in fig. S13A. More bistable units can be connected with feedback interaction. Fig. S14 shows three bistable units connected in series, where each unit can independently switch between two stable states in response to external light. We observed five stable configurations in the system, due to the mutual interaction between adjacent units. The transition dynamics and positioning of these five states under varying light intensities can be found in fig. S14C, D.

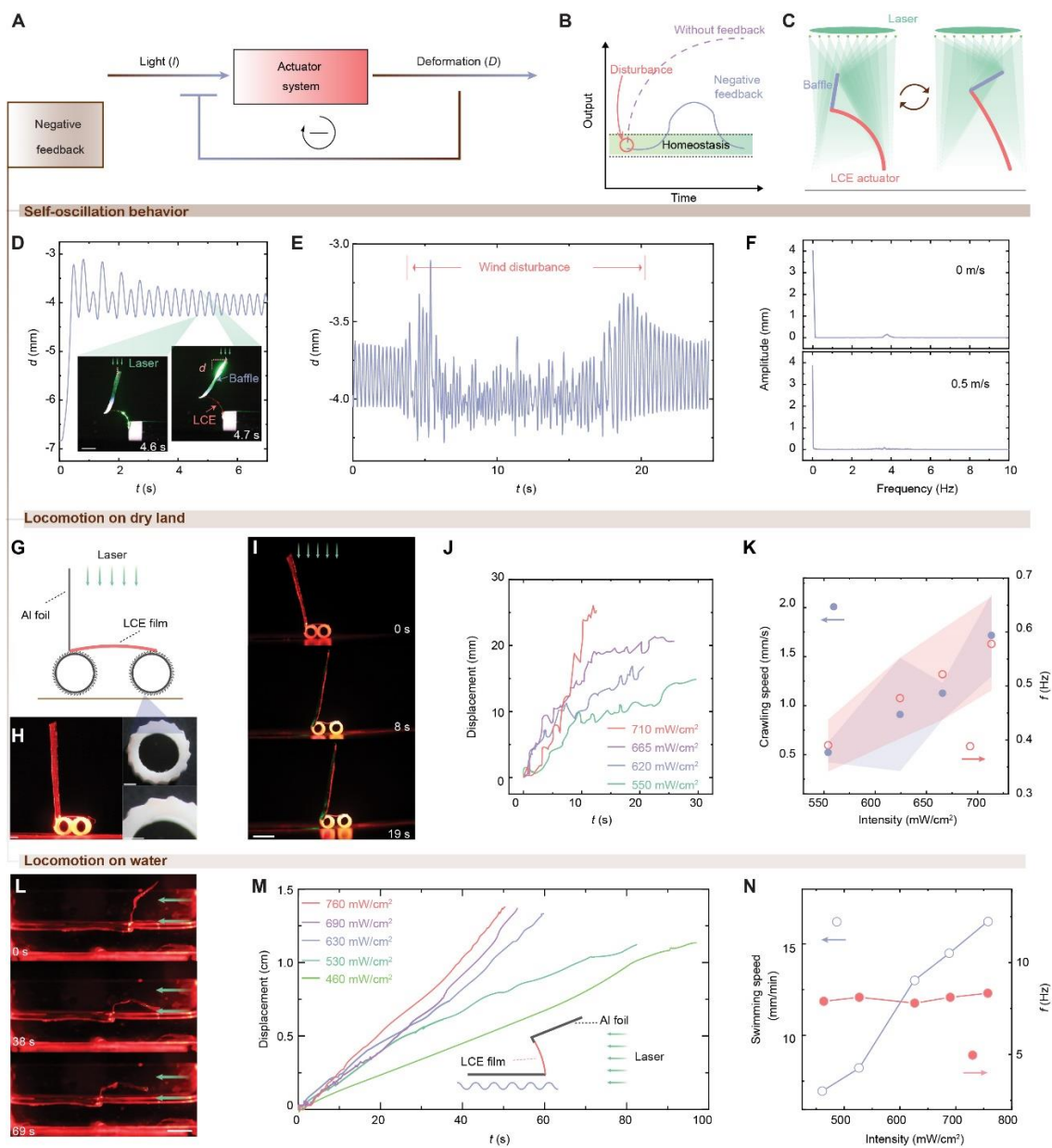

**Figure 3. Negative feedback in light-response actuators.**
(A) The schematics of the mechanism of negative feedback in the light-response actuator. (B) The schematics of homeostasis induced by negative feedback. (C) Schematics of a baffle-actuator system with negative feedback. (D) Displacement data of the baffle tip under a constant light field. Inset: snapshots of the self-oscillation process. Scale bar: 1 cm. Light intensity: 160 mW/cm$^2$. (E) Oscillation data of the light-response actuator upon wind disturbance. Light intensity: 160 mW/cm$^2$. Wind speed: 0.5 m/s. (F) Fourier transform of oscillation data upon different wind speed disturbances. Light intensity: 160 mW/cm$^2$. (G) The schematics of walking the robot controlled by negative feedback. (H) Left: photographs of a walking robot controlled by negative feedback; Right shows the microscopic image of the gear structure on the surface of wheel. Scale bars: 2.5 mm. (I) Snapshot images of the crawling behavior of the robot upon non-modulated light illumination from the top. Light intensity: 620 mW/cm$^2$. Scale bar: 1 cm. (J) Horizontal displacement of the crawling robot upon different light fuel intensities. (K) The crawling speed and the oscillation frequency $f$ of the crawling robot vs. light intensity. The error bars are displayed as mean values +/- standard deviation (n = 3). (L) Snapshot photos of a self-oscillating boat moving on the water surface. Light is irradiating horizontally from right to left with an intensity of 630 mW/cm$^2$. Scale bar: 1.5 cm. (M) Horizontal displacement of the floating boat upon light excitation with different light intensities. (N)

The mean swimming speed and mean frequency *f* of the floating boat vs. light intensity. The wavelength of light: 532 nm.

Furthermore, we illustrate the negative feedback mechanism of the light-responsive actuator in Fig. 3A. Light induces deformation of the actuator, which in turn dampens the effect of light on the actuator. As a result, the system moves around a certain steady point. This mechanism forms the basis of homeostasis, as depicted in Fig. 3B. Initially, the system stabilizes within a certain range, however, an external disturbance can cause the system to deviate from this stable range. In such cases, the negative feedback mechanism gradually guides the system back into the stable range. This kind of feedback loop plays a crucial role in maintaining the stability and robustness of the system against external perturbations. Fig. 3C illustrates the design of a negative feedback by using the same baffle and actuator as the positive one. In this case, the baffle is fixed at the end of the actuator, allowing a large portion of the incident light to pass through and reach the material. Once the actuator bends under light illumination, the baffle starts to block most of the light, thus functioning as negative feedback, $I \rightarrow D \dashv I$. Here, $D$ and $I$ refer to the deformation of the actuator and absorbed incident light energy, respectively. Regulated by such a negative feedback mechanism, the baffle-actuator exhibits self-oscillation behavior, as schematically explained in fig. S15. Under continuous illumination, the strip bends to the direction of the incident light, while the baffle at the tip of the strip obstructs the light. Following relaxation and cooling in darkness, the strip unbends, exposing itself to the light beam again, initiating a new cycle of motion. The self-oscillation is observed above the threshold of excitation intensity, as photographically shown in Fig. 3D and movie S3. The system shows a self-oscillation frequency of 3.8 Hz for an actuator with a length of 1.2 cm. The oscillation frequency depends on the sample dimension but remains independent of incident light intensity. Details of the correlation between oscillation amplitude, frequency, and actuator length, can be found in fig. S16. Applying the fast Fourier transform to the oscillation curve in the time domain can reveal more detailed information. As shown in Fig. 3E and 3F, figs. S17 and S18, the spectrum shows that once the characteristic length of 1.2 cm of the system is determined, within a reasonable range, no matter how the light intensity and the strength of the external disturbance are changed, the first harmonic component of the system remains steady, only the amplitude of this component changes. More importantly, during this process, regardless of the disturbance intensity, the average value of the system's direct component (tip displacement) remains stable at around 4 mm. This indicates that the system maintains its stability through spontaneous adjustment during receiving and dissipating input energy, counteracting the instability caused by external disturbances. This parallels the important biological phenomenon of homeostasis. Moreover, energy dissipation is accompanied by energy storage and utilization in order to maintain the system's steady state. Part of the energy is dissipated in the form of heat, while another portion of the energy is used to overcome friction or air resistance (*32*). We attempt to utilize the dissipated second portion of energy to perform useful work, thereby achieving robotic functions, *i.e.,* locomotion in both terrestrial and aquatic environments.

The first design aims to achieve locomotion on dry land, as shown in Fig. 3G and 3H, a negative feedback baffle-actuator is fixed on top of two cylindrical wheels, whose surfaces are 3D printed with ruler gratings to provide friction bias. The fabrication is detailed in fig. S19 and Materials and Methods Section. Self-oscillation induces cyclic slipping of the wheels on a flat surface. The friction bias enables net transportation of the structure towards the direction of lower friction, as shown by the snapshots in Fig. 3I and translocation data in Fig. 3J (also see movie S4). The crawling speed increases from 0.5 mm s$^{-1}$ to 1.7 mm s$^{-1}$ with the intensity risen from 550 mW/cm² to 710 mW/cm². Meanwhile, the oscillation frequency varies from 0.37 to 0.59 Hz (Fig. 3K).

The second design focuses on the realization of locomotion on the water surface by attaching the self-oscillating baffle-actuator to a floating boat. Details of fabrication are in fig. S19 and Materials and Methods Section. Upon the horizontal light beam excitation, the self-oscillating baffle generates wind flow through cyclic motion. The aerodynamic thrust propels the 23 mg boat away from the light source on the water surface. See the snapshots in Fig. 3L, movie S5, and mass displacement data in Fig. 3M. The swimming velocity increases along excitation intensity, while the oscillation frequencies remain constant at around 8 Hz being independent from light intensity (Fig. 3N). Swimming velocity shows a positive correlation to the oscillating amplitude (fig. S20), indicating an oscillation-induced air thrust mechanism responsible for the swimming motion. Worthy to be noted that, there is a distinction in frequency-intensity relation between the crawler and swimmer: the former varies in oscillation frequency when adjusting the incident intensity. We attribute this difference to the external friction force that influences the effective actuator length and in turn affects the negative feedback process during the crawling motion.

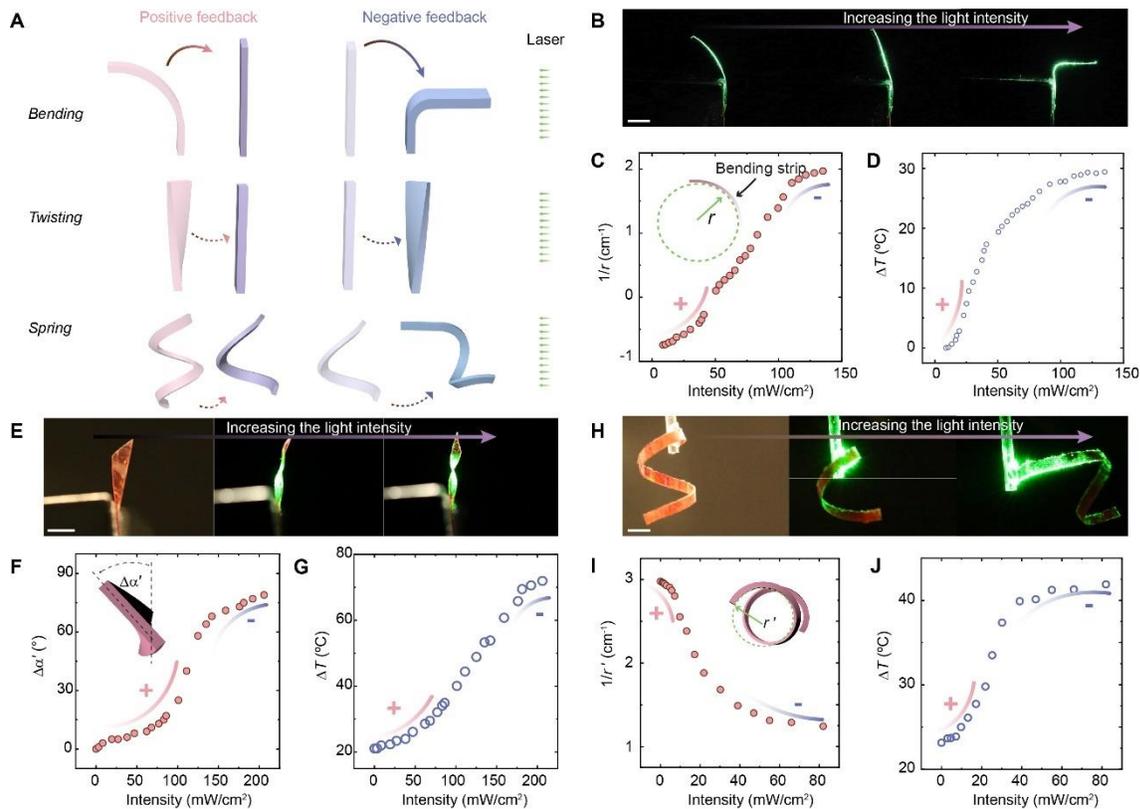

**Figure 4. The ubiquity of feedback in photo-mechanical matters.**
(A) Schematic drawing of the positive and negative feedback mechanism behind three deformation modes in light-responsive materials: bending, twisting, and spring-like deformation. (B) The optical photos of a bending strip actuated by an incident light field (horizontal direction, from right to left). Light intensity: 0 mW/cm² (left photo), 53 mW/cm² (middle photo) and 120 mW/cm² (right photo). (C) Deformation (the strip curvature, $1/r$, $r$ is the radius of the arc) of the bending strip at the thermal equilibrium upon the increase of illumination intensity. (D) Temperature elevation $\Delta T$ of the bending strip at the thermal equilibrium upon the increase of illumination intensity. (E) The optical photos of a twisting strip actuated by an incident light field (horizontal direction, from right to left). Light intensity: 0 mW/cm² (left photo), 100 mW/cm² (middle photo) and 200 mW/cm² (right photo). (F) Deformation (the change of twisting angle $\Delta\alpha'$) of the twisting strip at the thermal equilibrium upon the increase of illumination intensity. (G) Temperature increases of the twisting strip at the thermal equilibrium with increase of illumination intensity. (H) The optical photos of a spring actuator actuated by an incident light field (horizontal direction, from right to left). Light intensity: 0 mW/cm$^2$ (left photo), 40 mW/cm$^2$ (middle photo) and 80 mW/cm$^2$ (right photo). (I) Deformation (the spring curvature, $1/r´$, $r´$ is the radius of the circle of the spring actuator from a top view, see the inset) at the thermal equilibrium upon the increase of illumination intensity. (J) Temperature elevation of the spring actuator at the thermal equilibrium upon the increase of illumination intensity. All scale bars are 5 mm.

Incorporating feedback regulation into responsive materials is expected to develop and showcase advanced self-sustaining and autonomous behavior, which can interact with the external environment in a non-unidirectional way (Figures 2 and 3). Light-responsive materials, as a subclass of responsive materials, hold immense potential in robotics due to their wireless control capabilities and ease of miniaturization (*33-35*). The stimuli-responsiveness enables the remote control via a light field, while the microfabrication techniques (*36*) allow the creation of highly agile soft robots suitable for confined environments. However, implementing light-responsive materials in robotic systems is generally considered challenging. The reasons include the need for more precise focusing of light in space to accurately control light-driven micro-robotics for achieving multimodal motion as the size decreases. Moreover, the lightweight and flexible characteristics of the system lead to a loss of inherent stability under random disturbances.

Here, we attempt to raise a different perspective by bringing attention to feedback mechanisms existing in reported light-responsive soft matter actuators, an aspect largely overlooked in literature. Fig. 4A demonstrates three distinguishing modes of motion in light-responsive materials: bending, twisting, and spring-like deformation. Those experimental scenarios are selected due to their shared characteristics with the most commonly studied strip structures in the literature over the past decades. To achieve these fundamental deformations, the direction of light incidence is consistently fixed from right to left. Details regarding the preparation of light-responsive materials for twisting and spring-like deformations can be found in fig. S21. Fig. 4B shows the optical images of a light bending strip upon the incident of a horizontal light field, without any baffle attachment. The strip shows an initial bent geometry and undergoes deformation under light stimulation. Then, the strip gradually straightens when the light intensity increases. Fig. 4C shows an increase of slope in curvature ($1/r$)-intensity relation that, echoes with data shown in Fig. 1E, and is attributed to the positive feedback mechanisms. This is because, in this deformation step, the flattening leads to an increase of illuminated area, resulting in an enhancement of light energy absorption that further amplifies its own deformation. Subsequently, with a further increase in light intensity, the strip starts to bend in the opposite direction as the deformation begins to shield the light. In this deformation step, the curvature change reaches a maximum followed by a plateau, indicating the dominance of negative feedback mechanisms. The average temperature of the strip throughout the entire light-driven process is recorded (Fig. 4D and the thermal images in fig. S22), corresponding to the nonlinear light-response in deformation (Fig. 4C). Furthermore, light-responsive materials undergoing twisting and spring-like deformations without any attached baffles also exhibit characteristics of both positive and negative feedback under a light field stimulation (Figs. 4E and 4H). As the light intensity increases, the twisting strip gradually flattens, resulting in an increase in the illuminated area, as shown in Fig. 4F. The change of twisting angle $\Delta \alpha´$ with varying light intensity increases under the influence of the positive feedback mechanism. Subsequently, the twisting strip twists in the opposite direction, reducing the illuminated area, and demonstrating the characteristics of the negative feedback mechanism. Thermal imaging data (fig. S23) and Fig. 4G further support this process. For the spring-like deformation actuator, as the light intensity increases, its structure begins to unwind, rotating towards the direction of the light source, resulting in an increase in the illuminated area also (Fig. 4H, middle). As shown in Fig. 4I, the slope of curvature $1/r'$ change continuously increases (positive feedback). As the light intensity further increases, the actuator straightens out (Fig. 4H, right), the slope of curvature $1/r'$ change decreases until the curvature reaches a plateau (negative feedback). Thermal imaging data (fig. S24) and Fig. 4J similarly document the temperature trends under the dominance of both positive and negative feedback mechanisms. These experiments underscore the ubiquity of feedback mechanisms in photothermally responsive materials, emphasizing their fundamental role in regulating behavior and enabling complex adaptive responses to light stimulation and advanced functionalities.

**Discussion**

Recently, a growing field of autonomous materials systems is emerging, in which the researchers have reported the observation of continuous multi-mode motion upon non-modulated energy fields (*5, 10, 30, 37-39*). This negative feedback-based phenomenon has been documented across a wide array of material systems, highlighting its ubiquitous nature. As demonstrated, these motion mechanisms can serve as fundamental components for achieving complex and controllable functions in future micro-robots. We observed four main types of negative feedback mechanisms underlying the self-regulated behaviors. Firstly, the self-shadowing effect is exemplified by the above experiments (Fig. 1E and Figure 4). This effect refers to the phenomenon where the oscillation of an object is influenced by its own structural deformations, causing certain parts of the object to obstruct or shield other parts from external influences (*30, 40-47*). Worthy to be noted that, the change of transmittance due to the tunable photonic bandgap properties can also be used as a shadow to introduce feedback (*48*). In this way, periodic motion behavior and the homeostasis of physical quantities can be achieved. Secondly, negative feedback induces damping effects, which signifies a dissipative steady state where oscillatory motions gradually diminish over time. However, the damping introduces an intriguing capacity to sense the environmental influences, paving the way for novel functional avenues beyond the fundamental self-oscillation phenomenon (*21, 49, 50*). For instance, phototropism, where the system perceives stimulus direction and reacts, spontaneously and continuously adjusting its motion to strictly follow signal cues (*21, 51*). Other examples found in recent literature include self-regulation in deformability (*52*) and signal transduction *(10)*. Thirdly, EZ isomerization and crystallinity endow negative feedback and in turn can synergistically produce self-oscillation. EZ isomerization refers to the reversible transformation of a molecule between cis (Z) and trans (E) isomer configurations, the lifetime of which is dictated by the crystallinity. In the context of self-oscillation, the high content of Z-isomer induces a change of crystallinity that reduces the lifetime (thus concentration) of Z-isomer, endowing a negative feedback for self-oscillation in crystallinity and correlated material deformation (*53, 54*). Fourthly, zero elastic energy mode (ZEEM) behaviors. The net energy exchange over one cycle is essentially negligible in these ZEEMs, as exemplified by isometric over-angles, Möbius strips and elastic torus. The elastic energy gained in one part of the motion cycle is offset by energy losses in another part of the cycle, allowing the structure to continuously morph in shape upon the stimulated field, typically an energy field with gradient (*55-59*).

For positive feedback (*60, 61*), although less extensively studied, it holds intriguing potential. Several noteworthy examples stand out related to this topic, encompassing intriguing features such as flytrap-inspired grippers (*62*), adaptable irises (*63*), and the intriguing concept of self-enhanced light absorption driven by dye diffusion (*64*). These systems exhibit the ability to transition from one distinct state to another upon external stimuli – a phenomenon intimately entwined with the notion of memory (*65*). Crucially, it is essential to emphasize that this specific form of bistability is orchestrated by the external energy fields. Emerging through energy dissipation, this mechanism operates within a realm where originally no energy barriers stand between states, and the externally fueled feedback exerts control over the characteristics of the energy barriers. In stark contrast, mechanical bistable constructs (*31*), *e.g.,* snappers, are functionally based on the intrinsic material properties, which emerge from equilibrium states. They are characterized by the absence of any need for additional energy infusion; once established, they maintain their intrinsic states – a fundamental aspect of the system's behavior that remains unalterable post-construction.

In conclusion, by adjusting the position of the baffle on the photo-deformable materials, it becomes possible to create both positive and negative feedback mechanisms in LCE, which presents a promising prospect in microrobots. Under constant light excitation, multi-stable structures associated with positive feedback can perform the geometrical transition between mechanically stable states upon external triggers, where the energy barrier is dictated by the dissipative energy provided by a light fuel. It allows them to generate a large locomotion response from a tiny trigger, and once they reach a stable state, other moderate disturbances will not destroy its stability. Conversely, incorporating negative feedback for self-oscillation in a matrix of robots showcases homeostasis to attain the system's stability and perform useful work for locomotion on dry land or water surface. We further discussed the wide presence of positive and negative feedback mechanisms in light-responsive materials with different deformation modes and summarized the feedback mechanisms in the typical robotic matter examples from the literature. This work provides new insights into feedback mechanisms in light-responsive materials, creating significant opportunities for self-regulating soft matter robotics.

## Materials and Methods
### Materials
1,4-Bis-[4-(6-acryloyloxyhexyloxy)benzoyloxy]-2-methylbenzene (99%, RM 82) was purchased from SYNTHON Chemicals. 4-aminobutyric acid (99%) and dodecylamine (97%) were purchased from TCI. Disperse Red 1(95%) was purchased from Merck. All chemicals were used as received. All chemicals were used as received.

### Sample fabrication
To prepare liquid crystal cells, two coated glass substrates were bonded together. One substrate was coated with polyvinyl alcohol (PVA, 5 wt% in water, 3000 RPM, 1 min) and baked at 90 °C for 10 minutes before unidirectionally rubbing to achieve uniaxial alignment. The other substrate was coated with polyimide (PI, 3000 RPM, 1 min), followed by baking at 180 °C for 20 minutes to achieve homeotropic alignment. To determine the thickness of the liquid crystal elastomer (LCE) film, a gap was introduced between the two glass slides by introducing 100 μm microspheres (Thermo Scientific). The liquid crystal mixture containing 0.56 mmol RM 82, 0.2 mmol 4-aminobutyric acid, 0.2 mmol dodecylamine, and 2 wt.% 2,2-dimethoxy-2-phenylacetophenone (Irgacure 651) was melted at 100 °C and infiltrated into the cell via capillary at 100 °C and kept for 10 min. After cooling to 65 °C (1 °C/min), the cells were stored in an oven at 63 °C for 24 h to conduct the aza-Michael addition reaction (oligomerization). Then, the samples were irradiated with UV light (365 nm, 180 mW/cm$^2$, 10 min) for polymerization. Finally, the cells were opened with a razor blade. 1 mg of Disperse Red 1 was spread onto the surface of the samples and diffused into the elastomer on a hot plate (100 °C for 10 min).

### Fabrication of LCE actuator with baffle (Figure 1)
A paper strip (1.2 cm × 0.2 cm) was attached to one end of a 7 mm long LCE strip. On the opposite end of the paper strip, a rectangle shape aluminum foil was vertically adhered as the baffle. By adjusting the incident angle between the baffle and LCE strip, a deform-to-cut-on (positive feedback) or deform-to-cut-off (negative feedback) effect is enabled. Details of the assembly are in fig. S5.

### Fabrication of walker based on negative feedback
By utilizing an LCE strip (1 cm × 0.2 cm × 100 um), two wheels with the same serrated direction were interconnected. An aluminum foil (2 cm × 0.9 cm) was affixed to one end of the LCE strip, positioned on the top of the wheel. See fig. S19 for the details of the assembly.

### Fabrication of swimming boat based on negative feedback
An aluminum foil (2.3 cm × 1.7cm) was used as a floating boat, the end of which was attached to a vertical LCE strip (0.89 cm × 0.3 cm). Another aluminum foil (0.98 cm × 1.3 cm) was adhered to the free-moving end of the LCE to function as a baffle and aerodynamic propeller. Details of the assembly can be found in fig. S19.

### Energy barrier between bistable state
A thin fiber is used to push the aluminum film for state switching, the bending angle of the fiber correlates with the force applied perpendicularly to the tip. The energy barrier can be obtained by integrating the displacement of the aluminum film with the imposing force.

**Acknowledgments**

We acknowledge funding from Academy of Finland (Postdoctoral Researcher No. 331015 to H. Zhang, Research Fellow No. 340263 to H. Zeng). J. Y. acknowledges the support from China Scholarship Council (CSC). H. Zeng gratefully acknowledges the financial support of the European Research Council (Starting Grant project ONLINE, No. 101076207). H.P. acknowledges financial support from Academy of Finland Center of Excellence in Life-Inspired Hybrid Materials – LIBER (No. 346108).

**Author contributions:** H. Zeng conceived and supervised the project; J. Y. performed experiments with help from H. G., Z. D. and H. P.; J. Y., H.P. and H. Zeng wrote the manuscript taking the inputs from H. Zhang, H. P. and W. S.; All the authors discussed and contributed to the project. The feedback concept was developed by H. Zeng together with H. Zhang and H. P.

**Competing interests:** All other authors declare they have no competing interests.

**Data and materials availability:** All data are available in the main text or the supplementary materials.